\documentclass[openacc]{rstransa}

\usepackage{algorithm}
\usepackage{algpseudocode}
\usepackage{multicol}
\usepackage{subcaption}
\usepackage{underscore}
\usepackage{appendix}
\usepackage{chngcntr}

\titlehead{Research}

\begin{document}

\title{Bayesian Symbolic Regression via Posterior Sampling}

\author{
    G. F. Bomarito$^{1}$ and P. E. Leser$^{1}$}

\address{$^{1}$NASA Langley Research Center, Hampton, VA, USA}

\subject{Artificial Intelligence, Applied Mathematics, Statistics, Software}


\corres{G. F. Bomarito\\
    \email{geoffrey.f.bomarito@nasa.gov}}

\begin{abstract}
    Symbolic regression is a powerful tool for discovering governing equations directly from data, but its sensitivity to noise hinders its broader application. This paper introduces a Sequential Monte Carlo (SMC) framework for Bayesian symbolic regression that approximates the posterior distribution over symbolic expressions, enhancing robustness and enabling uncertainty quantification for symbolic regression in the presence of noise. Differing from traditional genetic programming approaches, the SMC-based algorithm combines probabilistic selection, adaptive tempering, and the use of normalized marginal likelihood to efficiently explore the search space of symbolic expressions, yielding parsimonious expressions with improved generalization. When compared to standard genetic programming baselines, the proposed method better deals with challenging, noisy benchmark datasets. The reduced tendency to overfit and enhanced ability to discover accurate and interpretable equations paves the way for more robust symbolic regression in scientific discovery and engineering design applications.
\end{abstract}

\begin{fmtext}
    \section{Introduction}

Symbolic regression (SR) stands out as a machine learning technique that promises compact, human-interpretable representations and the potential to uncover valuable physical insights\cite{kozaGeneticProgrammingProgramming1992,kronbergerSymbolicRegression2024}. Unlike traditional regression methods that require a pre-defined model structure, SR aims to discover the underlying mathematical expression that best describes relationships within a dataset. This capability has led to its successful application in diverse fields, including the identification of physical laws \cite{schmidtDistillingFreeFormNatural2009,hillsAlgorithmDiscoveringLagrangians2015,udrescuAIFeynmanPhysicsinspired2020,ohInherentlyInterpretableMachine2024}, discovery of governing equations \cite{bruntonDiscoveringGoverningEquations2016,cranmerDiscoveringSymbolicModels2020,bomaritoDevelopmentInterpretableDatadriven2021,cranmerInterpretableMachineLearning2023,birkyGeneralizingGursonModel2023,kronbergerSymbolicRegression2024,russeilMultiviewSymbolicRegression2024}, and the development of predictive regressors \cite{kozaGeneticProgrammingProgramming1992,lacavaFlexibleSymbolicRegression2023,kronbergerSymbolicRegression2024,merrellStressIntensityFactor2024,bartlettExhaustiveSymbolicRegression2024}. The appeal of SR lies in its ability to automatically learn both the form and parameters of a model directly from data, offering a powerful tool for scientific discovery and engineering design.
\end{fmtext}

\maketitle

A significant challenge hindering the more widespread adoption of SR is its sensitivity to noise and limited data availability. These are often the very scenarios where SR's unique ability to extract interpretable relationships from data would be most beneficial.  Recent benchmark studies have demonstrated that the effectiveness of many SR algorithms on system identification and predictive regression tasks is severely reduced when even nominal levels of noise are introduced into training datasets \cite{defrancaSRBenchPrincipledBenchmarking2024,lacavaSRBENCHResults2022}.  Thus, the propensity to overfit noisy data, coupled with the vast search space of possible symbolic expressions, makes the robustness of SR a critical area for improvement.

In this work, we pursue robust SR through the approximation of the distribution representing the probability of symbolic expressions given our observed dataset: \emph{i.e.,} the Bayesian posterior distribution over symbolic expressions. This approach aims to produce a family of symbolic expressions, each associated with a probability density reflecting its relative likelihood given the observed data. By characterizing the posterior distribution, we ultimately obtain (1) a maximum \emph{a posteriori} (MAP) expression, representing the most probable (\emph{i.e.}, maximally predictive) expression, and (2) the means to quantify and express uncertainty in predictions, both in the form of the expression and in its parameters. This uncertainty quantification is crucial for assessing the robustness and reliability of SR models, especially in noisy environments.

To efficiently explore the vast space of possible symbolic expressions and approximate the Bayesian posterior, we leverage the power of Sequential Monte Carlo (SMC) \cite{delmoralSequentialMonteCarlo2006}. SMC provides a framework for iteratively refining a population of symbolic expressions, guiding the search towards regions of non-zero posterior probability density. By maintaining a diverse set of candidate models and adapting the population based on evidence from the data, SMC enables us to effectively handle the challenges posed by noise and limited data. Furthermore, the population-based nature of SMC readily allows for the estimation of multimodal posteriors, which are common in SR applications \cite{leserComparingMethodsEstimating2024}.

In related works \cite{guimera2020bayesian,jinBayesianSymbolicRegression2020}, Bayesian SR has been performed using Markov Chain Monte Carlo (MCMC) \cite{smithUncertainty2024} for posterior sampling; however, MCMC has a few notable drawbacks in this application.  First, MCMC is less scalable than SMC, requiring long chains of calculations that must be performed serially.  Secondly, MCMC can struggle to efficiently explore complex and high-dimensional domains such as that of symbolic expressions; this is particularly true when the posterior distribution is multi-modal. The difficulty of search could potentially be addressed by carefully tuning a transition kernel to a given problem, applying parallel tempering \cite{guimera2020bayesian}, or by using more advanced, adaptive MCMC algorithms; however, methodology for effective adaptation in the domain of SR is still lacking. On the other hand, the effectiveness of population-based algorithms for SR is underscored by the fact that genetic-programming-based SR (GPSR) methods currently achieve state-of-the-art performance on many benchmark datasets \cite{lacavaContemporarySymbolicRegression2021,defrancaSRBenchPrincipledBenchmarking2024}. Thus, we posit that because SMC is a population-based method, it provides a natural and effective framework for performing Bayesian inference in the complex landscape of symbolic expressions.

Other Bayesian approaches, specifically Bayesian model selection, have previously been incorporated into GPSR \cite{bomaritoBayesianModelSelection2022,bomaritoAutomatedLearningInterpretable2023,bartlettPriorsSymbolicRegression2023}.  These works have illustrated benefits to GPSR including reducing bloat, increasing generalizabity, and introduction of physical knowledge through definition of a prior.  Here, we aim to expand on these works by maintaining these benefits while incorporating a principled application of Bayesian inference. This addition ensures that our final population of expressions converges to an approximation of the Bayesian posterior.

This paper introduces a novel SMC-based algorithm for Bayesian SR, aimed at addressing the limitations of existing MCMC-based methods and expanding on previous Bayesian GPSR efforts. Our primary contributions include:
\vspace{-1em}
\begin{enumerate}
    \item The development of a robust SMC framework for effectively exploring the space of symbolic expressions and approximating the Bayesian posterior,
    \item A thorough evaluation on benchmark datasets, highlighting the superior performance (better accuracy and generalizability) of the proposed method over GPSR in noisy environments,
    \item A discussion of likely reasons for improved performance with supporting analysis.
    \item Open-source software for the proposed approach: \url{github.com/nasa/pysips}
\end{enumerate}
\vspace{-1em}
The remainder of this paper details the methodological details of our approach (Section \ref{sec:methods}), including the specific SMC implementation used and comparison with other methods. Subsequently (Section \ref{sec:results}), we present experimental results demonstrating the efficacy of our method on a range of benchmark datasets, specifically highlighting its robustness to noise. Finally, in Sections \ref{sec:discussion} and \ref{sec:conclusion}, we discuss the implications and nuances associated with these findings and outline potential directions for future research.

\section{Methods}
\label{sec:methods}

The Bayesian approach to SR, henceforth referred to as Baysian SR, involves estimating the joint posterior distribution over models, $\mathcal{M}$, and parameters, $\pmb{\theta}$, given observed data, $\mathcal{D}$:
\begin{align}\label{eq:posteriorOnParamtersAndModels}
    \underbrace{\pi(\pmb{\theta}, \mathcal{M} | \mathcal{D})}_{\text{Joint Posterior}} = \frac{\overbrace{\pi(\mathcal{D} | \pmb{\theta}, \mathcal{M})}^{\text{Likelihood}}\pi(\pmb{\theta}| \mathcal{M})\pi(\mathcal{M})}{\pi(\mathcal{D})}.
\end{align}
Here, $\mathcal{M}$ is a function of $\pmb{\theta} \in \mathbb{R}^{N_\theta}$ and independent variables $\mathbf{x} \in \mathbb{R}^{N_x}$, \emph{i.e.}, $\mathcal{M}(\mathbf{x}, \pmb{\theta}) \approx y$, and the dataset $\mathcal{D}$ contains $N_d$ observations of $(\mathbf{x}, y)$ pairs. The domain of $\mathcal{M}$ is discrete while the domain of $\pmb{\theta}$ is continuous with dimension $N_\theta$ varying for a given $\mathcal{M}$. Exploration of the trans-dimensional space of models and parameters is difficult. Jin et al.\cite{jinBayesianSymbolicRegression2020} used reversible jump MCMC to estimate an approximation of $\pi(\pmb{\theta}, \mathcal{M} | \mathcal{D})$; however, their method still required a number of simplifications and constraints on the model structure to address inefficient sampling.

Given that the ultimate goal in SR is to identify models matching the data, \eqref{eq:posteriorOnParamtersAndModels} can be marginalized over the parameter dimension, resulting in the posterior over models,
\begin{align}\label{eq:posteriorOnModels}
    \underbrace{\pi(\mathcal{M} | \mathcal{D})}_\text{Model Posterior} = \int_\mathbb{R^{N_\theta}} \pi(\pmb{\theta}, \mathcal{M} | \mathcal{D}) d\pmb{\theta} \quad \propto \quad \underbrace{\pi(\mathcal{M})}_\text{Model Prior} \underbrace{\int_\mathbb{R^{N_\theta}} \pi(\mathcal{D} | \pmb{\theta}, \mathcal{M})\pi(\pmb{\theta}| \mathcal{M})d\pmb{\theta}}_{\text{Marginal Likelihood}}.
\end{align}
The primary benefit of marginalization is that the model space can be explored without requiring trans-dimensional jumps in $\pmb{\theta}$, leading to more efficient approximation of the posterior with standard sampling algorithms (\emph{e.g.}, MCMC and SMC). The drawback of the marginalization is that the integral on the right hand side, referred to as the marginal likelihood, must be computed for every proposed $\mathcal{M}$, resulting in a computationally-intensive double loop. Each iteration of the inner loop is equivalent to solving for the normalizing constant of the more classic, fixed-model Bayesian inference problem.  In other words, we gain efficiency in navigation between models at the cost of higher computational cost for each sampled model.

The efficient approximation of the marginal likelihood in a SR framework is the subject of previous works \cite{bomaritoBayesianModelSelection2022,bomaritoAutomatedLearningInterpretable2023,bartlettPriorsSymbolicRegression2023}. It was noted that little prior information on $\pmb{\theta}$ is available in SR since the model form is unknown \emph{a priori}. Therefore, the prior is often chosen to be an uninformative, improper uniform distribution, \emph{i.e.,} $\pi(\pmb{\theta}| \mathcal{M}) \propto 1$, which results in an indeterminate constant appearing in the marginal likelihood. Originally proposed in \cite{ohaganFractionalBayesFactors1995}, the use of a normalized marginal likelihood (NML) has been used to alleviate this issue in SR\cite{bomaritoBayesianModelSelection2022,bomaritoAutomatedLearningInterpretable2023,bartlettPriorsSymbolicRegression2023,leserComparingMethodsEstimating2024},
\begin{align}
    q = \frac{\int \pi(\pmb{\theta} | \mathcal{D}, \mathcal{M})\pi(\pmb{\theta}| \mathcal{M})}{\int \pi(\pmb{\theta} | \mathcal{D}, \mathcal{M})^\gamma\pi(\pmb{\theta}| \mathcal{M})},
\end{align}
where the empirically motivated choice of $\gamma=1/\sqrt{N_d}$ is used \cite{ohaganFractionalBayesFactors1995}.

The NML can be estimated directly with SMC \cite{bomaritoBayesianModelSelection2022,bomaritoAutomatedLearningInterpretable2023} or approximated using the Laplace approximation \cite{bartlettPriorsSymbolicRegression2023},
\begin{align}\label{eq:laplace}
    \hat{q} = \gamma^{N_\theta/2}\pi(\mathcal{D}| \pmb{\theta}^*, \mathcal{M})^{(1-\gamma)}\approx q.
\end{align}
Here, $\pmb{\theta}^*$ is the maximum \emph{a posteriori} estimate, which is equivalent to the maximum likelihood estimate given the improper uniform prior.  In practice, $\pmb{\theta}^*$ is estimated using a gradient-based optimization. The Laplace approximation is orders of magnitude faster than using SMC to estimate NML but is less accurate for many common SR models \cite{leserComparingMethodsEstimating2024}. We use the Laplace approximation to estimate $q$ in this work as it was assumed that an inner loop implementation of SMC would be computationally intractable.

Given the NML estimator and a choice of prior over models, $\pi(\mathcal{M})$, MCMC or SMC can be used to draw samples from the posterior \eqref{eq:posteriorOnModels} as a means of performing Bayesian SR. In this work, we choose SMC for the following reasons: (1) MCMC requires a carefully tuned proposal or can suffer from very low acceptance rates whereas SMC, a global, population-based algorithm, is generally more robust to proposal selection; (2) MCMC has strict ergodicity requirements to ensure that the stationary distribution of the Markov chain is equivalent to $\pi(\mathcal{M}|\mathcal{D})$, whereas SMC can relax these requirements; and (3) MCMC is fundamentally serial whereas SMC can be parallelized to improve computational efficiency.

\subsection{Posterior Sampling with Sequential Monte Carlo}


SMC-based SR (SMC-SR) works by evolving a population of weighted symbolic expressions, $\mathcal{P} = \left\{ \mathcal{M}_1, \dots, \mathcal{M}_{N_P} \right\} $, through a series of target distributions using sequential importance sampling and resampling.  Using a process called likelihood tempering, the $t^{\mathrm{th}}$ target distribution is defined as
\begin{equation}
\pi_t(\mathcal{M}|\mathcal{D}) \propto \pi(\mathcal{M}) \left[\int_\mathbb{R^{N_\theta}} \pi(\mathcal{D} | \pmb{\theta}, \mathcal{M})\pi(\pmb{\theta}| \mathcal{M})d\pmb{\theta}\right]^{\phi_t}\propto\pi(\mathcal{M})\hat{q}(\mathcal{M})^{\phi_t}
\end{equation}
with $0=\phi_0 < \ldots < \phi_t < \ldots < \phi_T = 1$ such that target distributions transition smoothly from the prior to the posterior. In this way, the initial population can be sampled directly from $\pi(\mathcal{M})$, which is known, where the $i^{\mathrm{th}}$ expression is assigned normalized weight $W_i=1/{N_P}$.

At each step, $\phi_t$ is updated adaptively \cite{buchholzAdaptiveTuningHamiltonian2021} and the expressions are reweighted according to the new target $\pi_t$ using importance sampling,
\begin{equation}\label{eq:reweight}
W^{(t)}_i = \frac{W_i^{(t-1)} \hat{q}(\mathcal{M}_i)^{\Delta{\phi_t}}} {\sum_{i=1}^{N_P} W_i^{(t-1)} \hat{q}(\mathcal{M}_i)^{\Delta{\phi_t}}}.
\end{equation}
The update $\Delta{\phi_t} = \phi_{t} - \phi_t $ is chosen using a bisection algorithm to maintain a user-specified effective sample size (ESS) where ESS is defined as $1\big/\sum_{i=1}^{N_P}{(W_i)^2}$. The ESS can vary between $1$ and $N_P$ and is an estimate of the number of expressions with non-negligible weights.

If the originally-sampled population of expressions is held fixed as $\phi_t \rightarrow 1$, it is likely that the weights will become degenerate (\emph{i.e.}, ESS is low and many expressions have $W_i\approx0$ because they lie far from the region of high posterior density). In the context of SR, this manifests as a drastic reduction in the number of highly-fit equations in the population. To avoid degeneracy, two strategies are implemented in the SMC algorithm. First, the expressions are reweighted according to the new target and resampled with replacement. During resampling, equations with large $W_i$ tend to be replicated and equations with low $W_i$ tend to be removed from the population. Stratified resampling \cite{holResampling2006} was used in this work to encourage expression diversity, but resampling iterations will reduce the number of unique equations in the population. To counteract this, the second strategy invovles short runs of a forward MCMC kernel to produce a rejuvenated population approximately distributed according to the current target, $\mathcal{P} \sim \pi_{t}(\mathcal{M}|\mathcal{D})$. In other words, the duplicated expressions are independently modified with MCMC. The combination of resampling and MCMC moves promotes diversity and global exploration of posterior modes. 
For more general information on SMC, see \cite{delmoralSequentialMonteCarlo2006}.

Algorithm \ref{alg:smcSR} provides a summary of the proposed SMC-SR algorithm and contrasts it with a variant of GPSR, which is summarized in Algorithm \ref{alg:gpsr}. Here GPSR with deterministic crowding selection \cite{mengshoel2008crowding} is chosen because its brood-style selection is especially similar with SMC-SR. It is clear from the comparison that the primary differences are the SMC-specific steps (\emph{e.g.}, iterating through target distributions, reweighting, resampling, and the MCMC forward kernel) and the definition of the acceptance probability, $\alpha$.  For the SMC approach, the acceptance probability is the classic random walk Metropolis step whereas traditional GPSR performs a binary accept/reject based on whether fitness has improved or not. The key ingredients for the MCMC kernel are $\alpha$ and proposal distribution, $h(\mathcal{P}'|\mathcal{P})$, used to generate offspring, $\mathcal{P}'$, from parents in $\mathcal{P}$: \emph{i.e.}, 
\begin{equation}\label{eq:proposalDistribution}
\mathcal{P}' \sim h(\mathcal{P}'|\mathcal{P})
\end{equation}
where $\mathcal{P}' = \left\{ \mathcal{M}'_1, \dots, \mathcal{M}'_{N_P} \right\}$. Standard GPSR variation operations of population-based crossover and mutation \cite{randallBingoCustomizableFramework2022} are used for the proposal, $h(\mathcal{P}'|\mathcal{P})$; details are included in the supplementary material (Appendix A).

\algrenewcommand{\algorithmicindent}{1em} 

\algdef{SE}[REPEATN]{RepeatN}{End}[1]{\algorithmicrepeat\ #1 \textbf{times}}{\algorithmicend}
\algdef{SE}[EVERYN]{EveryN}{End}[1]{\textbf{at every} #1 \textbf{repetition}}{\algorithmicend}

\setlength{\columnsep}{0pt}
\begin{figure*}[t] 
    \begin{multicols}{2}

        \begin{algorithm}[H]
            \caption{Sequential Monte Carlo\\\vspace{0.28em}}\label{alg:smcSR}
            \begin{algorithmic}[1]
                \State \textbf{given} a dataset, $\mathcal{D}$
                \State \textbf{define} $\hat{q}(\cdot)$ as the NML \eqref{eq:laplace} given $\mathcal{D}$
                \State Generate a set of symbolic expressions $\mathcal{P}$
                \State Evaluate $\hat{q}(\mathcal{M}_i)$ for $\mathcal{M}_i$ in $\mathcal{P}$
                \State Initialize $\phi_t=0$
                \State Set uniform weights ${W_i}=1/N_p$
                \Repeat
                \State Update $\phi_t$ \cite{buchholzAdaptiveTuningHamiltonian2021}
                \State Update weights $W$ \eqref{eq:reweight}
                \State Resample $\mathcal{P}$ with replacement \cite{holResampling2006}
                \RepeatN{$N_{MCMC}$}
                \State Generate offspring $\mathcal{P}'$ of $\mathcal{P}$ \eqref{eq:proposalDistribution}
                \State Evaluate $\hat{q}(\mathcal{M}'_i)$ for $\mathcal{M}'_i$ in $\mathcal{P}'$
                \State Pair parent $\mathcal{M}_i$ with offspring $\mathcal{M}'_i$
                \State $\alpha = \min(1, (\hat{q}(\mathcal{M}'_i)/\hat{q}(\mathcal{M}_i))^{\phi_t})$
                \State Replace $\mathcal{M}_i$ with $\mathcal{M}'_i$ with probability $\alpha$
                \End
                \vskip 0.06\baselineskip
                \Until{$\phi_t = 1$}
            \end{algorithmic}
        \end{algorithm}

        \columnbreak

        \begin{algorithm}[H]
            \caption{Genetic Programming\\(deterministic crowding selection)}\label{alg:gpsr}
            \begin{algorithmic}
                \State \textbf{given} a dataset, $\mathcal{D}$
                \State \textbf{define} $e(\cdot)$ as a loss function given $\mathcal{D}$
                \State Generate a set of symbolic expressions $\mathcal{P}$
                \State Evaluate $e(\mathcal{M}_i)$ for $\mathcal{M}_i$ in $\mathcal{P}$
                \State $\cdot$
                \State $\cdot$
                \Repeat
                \State $\cdot$
                \State $\cdot$
                \State $\cdot$
                \State $\cdot$
                \State Generate offspring $\mathcal{P}'$ of $\mathcal{P}$  \eqref{eq:proposalDistribution}
                \State Evaluate $e(\mathcal{M}'_i)$ for $\mathcal{M}'_i$ in $\mathcal{P}'$
                \State Pair each parent $\mathcal{M}_i$ with its offspring $\mathcal{M}'_i$
                \State $\alpha = 1$ if $e(\mathcal{M}'_i) \leq e(\mathcal{M}_i)$ else $0$
                \State Replace $\mathcal{M}_i$ with $\mathcal{M}'_i$ with probability $\alpha$
                \State $\cdot$
                \Until{convergence or repetition limit}
            \end{algorithmic}
        \end{algorithm}

    \end{multicols}
\end{figure*}

It is worth noting that our proposal strategy is asymmetric, meaning $h(\mathcal{P}'|\mathcal{P})\ne h(\mathcal{P}|\mathcal{P}')$.  This can be detrimental to MCMC due to its strict ergodicity requirement. We instead rely on the fact that our MCMC kernel is implemented within the broader SMC algorithm where ergodicity improves sampling but is not strictly required as long as the kernel is effectively rejuvenating the population. The combination of global exploration, less sensitivity to proposal selection, relaxation of ergodicity requirements, and computational efficiency made SMC a natural choice over MCMC for this initial study. A more detailed comparison of SMC and MCMC for Bayesian SR is included in the supplementary material (Appendix B).

A number of additional choices were made that are briefly summarized here for repeatability. First, throughout this work it is assumed that noise in the data is independent and identically distributed (iid) such that, for a given model, $y_j=\mathcal{M}(\mathbf{x}_j, \pmb{\theta}) + \epsilon_j$ where $\epsilon_j \overset{\text{iid}}{\sim} N(0, \sigma)$ and $\sigma$ could be estimated as part of the inference. The prior over models is assumed to be uniform, \emph{i.e.}, $\pi(\mathcal{M})\propto1$. However, it should be noted that this was a choice and not a requirement of the method. Informative priors such as those proposed by Bartlett et al.\cite{bartlettPriorsSymbolicRegression2023} could be implemented, and simply impart a user's belief upon the likely model structure.
Here, expressions were randomly generated for the initial population $\mathcal{P}$ until a unique set of expressions of desired size was achieved.\footnote{Equality testing of expressions occurs after symbolic simplification and reorganization into a canonical form. However, these processes are imperfect and may not identify all symbolic equivalences.} Though this generation process likely imparts some nonuniformity, it is an approximation of samples of the uniform prior $\pi(\mathcal{M})\propto1$.

A handful of hyperparameters are required for the SMC algorithm. They are listed here and will remain constant for the remainder of this work (unless otherwise noted).  The total number of expressions in the population was $N_P=2000$. To ensure the forward kernel provided good mixing (generally measured as the percentage of final offspring that differ from the original parent, where higher percentages are better), $N_{MCMC}$ was set to 10. The target ESS was 1900, or 95\% of the population size. Higher percentages generally result in smaller $\phi_t$ updates and thus more iterations. No effort was made to optimize hyperparameters (though future work may find this fruitful), besides observing that relatively large values of $N_P$ are preferred to smaller sizes. Although larger populations impart a larger computational burden, we posit that larger population sizes allow for more expression diversity early on in sampling, which in turn leads to more efficient exploration of intermediate targets and, ultimately, the posterior.

\subsection{Demonstration}
\label{sec:demo}
The key aspects of SMC-SR are demonstrated here using a simple case study.  A dataset was generated with $y_j = \frac{1}{2\pi}e^{-\frac{(x_j)^2}{2}} + \epsilon_j$ where $x_j \sim \mathcal{U}(-3, 3)$ and $\epsilon_j \sim \mathcal{N}(0, 0.16)$. The dataset comprised 25 datapoints.  SMC-SR was applied to this dataset using the operators $[+, -, \times]$ in order to make the regression more difficult since SMC could quickly identify the true expression given an exponential operator.

\begin{figure}
    \centering
    \captionsetup[subfigure]{belowskip=-1pt}
    \begin{subfigure}[b]{0.49\textwidth}
        \centering
        \includegraphics{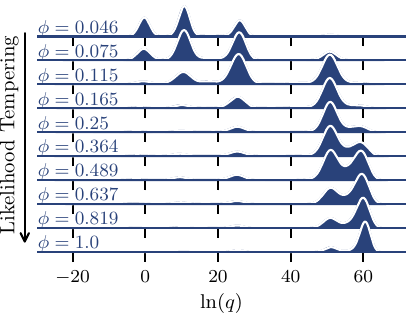}
        \caption{}
        \label{fig:demoNmll}
    \end{subfigure}
    \hfill
    \begin{subfigure}[b]{0.49\textwidth}
        \centering
        \includegraphics{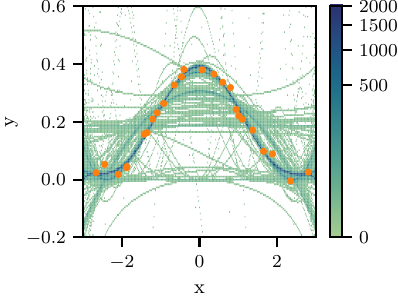}
        \caption{}
        \label{fig:demoY}
    \end{subfigure}
    \caption{Demonstration of SMC-SR. (a) The effect of likelihood-tempering on the distribution of NML. (b) The model posterior viewed as a histogram. Points indicate the training data.}
    \label{fig:demonstration}
\end{figure}

Due to difficulties visualizing models sampled from the posterior $\pi(\mathcal{M}|D)$, we instead show two alternatives in Figure \ref{fig:demonstration}.  In Figure \ref{fig:demoNmll} the distribution of log of NML, $\text{ln}(q)$, is shown which allows for the visualization of the posterior in a single dimension. The distribution is also shown for some intermediate target distributions (\emph{i.e.}, at intermediate $\phi_t$ values). The figure illustrates how likelihood-tempering in SMC corresponds to early exploration (larger spread in distribution at $\phi=0.046$) followed by gradual convergence toward a multimodal posterior. Figure \ref{fig:demoY} depicts the model posterior as a two-dimensional histogram in the space of the data.  Over most of the domain, there is very high, unimodal density in the areas between the training data points with histogram frequency of nearly 2000 (the population size used).  Areas with disagreement between models (which look more multimodal) indicate two things: an area where caution might be exercised when making predictions and also an area where additional training data might be especially insightful.

Note that Figure \ref{fig:demoY} is an effort to illustrate the posterior $\pi(\mathcal{M}|D)$, but it should not be confused with the joint posterior $\pi(\pmb{\theta}, \mathcal{M} | \mathcal{D})$ which is commonly used to create credible intervals.  However, the process is straightforward to approximate $\pi(\pmb{\theta}, \mathcal{M} | \mathcal{D})$ from our approximation of $\pi(\mathcal{M}|D)$: using the final $\mathcal{P}$ produced with SMC-SR and performing probabilistic calibrations of the $\pmb{\theta}$ in each model (using \emph{e.g.}, standard MCMC or SMC tools).

\section{Results}
\label{sec:results}

The efficacy of SMC-SR was quantified using a set of 12 benchmark problems.  These benchmark problems were modelled after equations seen in Richard Feynman's physics lectures \cite{udrescuAIFeynmanPhysicsinspired2020} and have been incorporated into the SR benchmarking suite SRBench \cite{lacavaContemporarySymbolicRegression2021,defrancaSRBenchPrincipledBenchmarking2024}.
Previous results on the Feynman datasets have illustrated that the addition of even 1\% noise into the datasets drastically reduces performance of most SR algorithms \cite{lacavaSRBENCHResults2022}; thus, noisy versions of the Feynman datasets represent a pertinent challenge for Bayesian SR methods.
The 12 specific datasets chosen for this work are chosen based on the most difficult subset of Feynman datasets \cite{matsubaraRethinkingSymbolicRegression2022} that are scheduled to be incorporated into the upcoming revision of  SRBench.\footnote{\url{https://github.com/cavalab/srbench/discussions/174\#discussioncomment-10285133}} Training on the datasets are repeated 20 times in this study.

The datasets each consist of 10,000 data points that are obtained by evaluating a known physical equation over a range of its input parameter space. The dimension of the input, $N_x$, ranged from 3 to 8. Training subsets with size $N_d=500$ data points were extracted, and the remaining data points were withheld as test sets.  Gaussian noise was added to the training set with standard deviation of 10\% of the magnitude of the dataset, \emph{i.e.}, $0.1||Y||$. The magnitude was calculated as $||Y|| = \text{median}(|y|)$ for the datasets because many contain data points very close to asymptotes that could skew other norms.  Normalized root mean squared error (NRMSE) will be used when discussing results; here the standard root mean squared error is normalized by $||Y||$.  Thus, a NRMSE of 0.1 corresponds to achieving an accuracy level comparable to the added noise.  We define here NRMSE-train and NRMSE-test as two metrics corresponding to NRMSE calculated using the training dataset and testing dataset, respectively.

SMC-SR is compared to 3 GPSR baselines.  The GPSR baselines have minimal variation from SMC-SR (see Algorithm \ref{alg:gpsr}) to make as direct a comparison as possible.  For instance, the population size in GPSR and SMC-SR are both set to a fixed value of $N_P=2000$.
The GPSR baselines vary based on their loss function and selection algorithm as indicated below:
\vspace{-1em}
\begin{center}
\begin{tabular}{c|c|c}
    & Loss Function & Selection Algorithm \\ \hline \hline
    GP-MSE & MSE & deterministic crowding (less aggressive)\\
    GP-NML & NML & deterministic crowding (less aggressive)\\
    GP-agg & MSE & tournament (more aggressive)\\
\end{tabular}
\end{center}
\vspace{-1em}
Since the SMC algorithm is adaptive, $T$, the total number of $\phi_t$ updates, is not fixed \emph{a priori}.  In order to preserve a consistent level of compute, the SMC-SR benchmarks are performed first and the GPSR benchmarks are performed subsequently with a number of generations equal to $T \times N_{MCMC}$.  Repetitions of the algorithms on the same dataset may or may not have the same number of SMC steps; hence, a normalized compute metric $[0-1]$ is used to indicate algorithm progress in further figures and discussions.

\begin{figure}
    \centering
    \captionsetup[subfigure]{belowskip=2pt}
    \begin{subfigure}[b]{0.49\textwidth}
        \centering
        \includegraphics{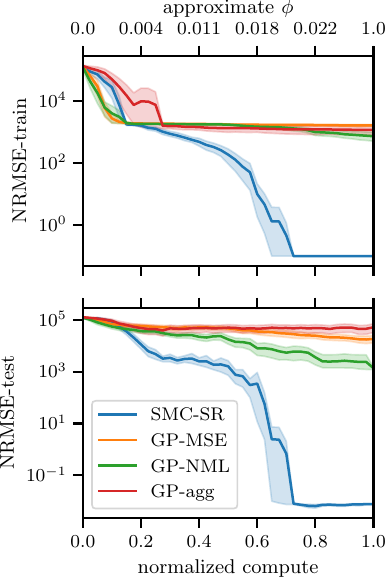}
        \caption{}
    \end{subfigure}
    \hfill
    \begin{subfigure}[b]{0.49\textwidth}
        \centering
        \includegraphics{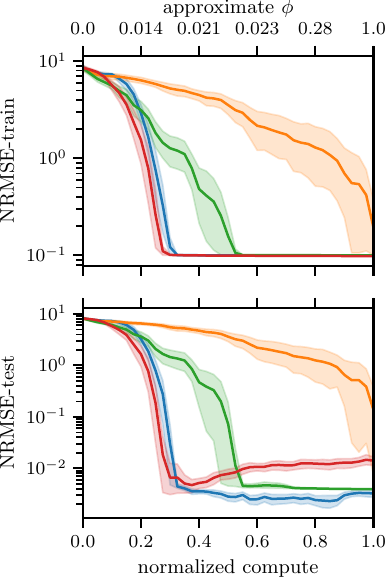}
        \caption{}
    \end{subfigure}
    \caption{Training results for Feynman dataset (a) I-32-17 and (b) II-36-38. Figures show the lowest (top) NRMSE-train and (bottom) NRMSE-test in the populations.}
    \label{fig:feynIndvError}
\end{figure}

The effectiveness of SMC-SR on two selected Feynman datasets is illustrated in Figure \ref{fig:feynIndvError}.  For dataset I-32-17, lower levels of NRMSE-train are achieved with SMC-SR than any GPSR method.  By comparing GP-MSE and GP-NML we can see that the addition of the regularization provided by the NML loss is partly responsible for the improved performance of SMC-SR. For dataset I-36-38, GP-agg achieves a low level of NRMSE-train quicker than SMC-SR, indicating that a high selection pressure (provided by resampling in SMC-SR and by tournament selection in GP-agg) leads to more rapid fitting of the training data.  However, the high values of NRMSE-test for GP-agg indicate that the higher selection pressure leads GP-agg to overfitting while SMC-SR does not.  This poor performance of GP-agg indicates that higher selection pressure cannot be solely the cause for improved performance in SMC-SR. The minimum NRMSE-test over the population $P$ is shown to illustrate the presence of highly-fit expressions in the population.  These highly-fit expressions are encountered more quickly with SMC-SR.  The practical ability to identify the highly-fit expressions without access to a large test dataset, especially with a population size of 2000 expressions, is discussed later in this section.

\begin{figure}
    \centering
    \captionsetup[subfigure]{belowskip=-1pt}
    \begin{subfigure}[b]{0.49\textwidth}
        \centering
        \includegraphics{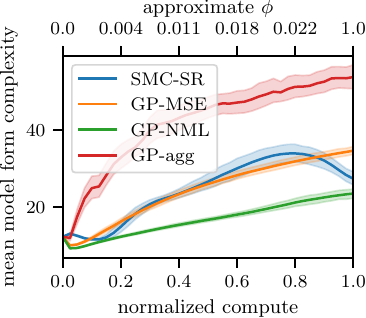}
        \caption{}
    \end{subfigure}
    \hfill
    \begin{subfigure}[b]{0.49\textwidth}
        \centering
        \includegraphics{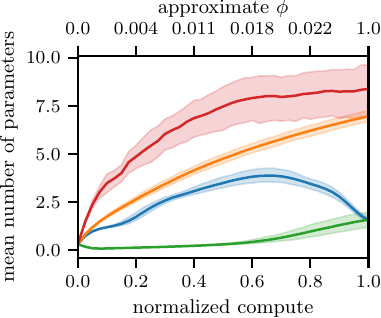}
        \caption{}
    \end{subfigure}
    \caption{Complexity of models for the Feynman dataset I-17-32 in terms of (a) model form and (b) number of parameters.}
    \label{fig:feynIndvComplecity}
\end{figure}

The effects of the SMC algorithm on number of parameters ($N_\theta$) and model form complexity (number of nodes needed to represent an expression as an acyclic graph) are illustrated in Figure \ref{fig:feynIndvComplecity}. Early on in SMC-SR, while in a state of exploration with $\phi \ll 1$ the complexity and number of parameters rise very quickly, at approximately the rate of GP-MSE.  But as the NML is tempered and $\phi\rightarrow1$, regularization appears to take hold and complexities and numbers of parameters approach that of GP-NML.  In other words, SMC-SR first explores and bloats but eventually identifies a region of high posterior density and homes in on those expressions in a way that reduces complexity. The rapid complexity growth for GP-agg illustrates another pitfall seen when blindly increasing selection pressure.

\begin{figure}
    \centering
    \includegraphics{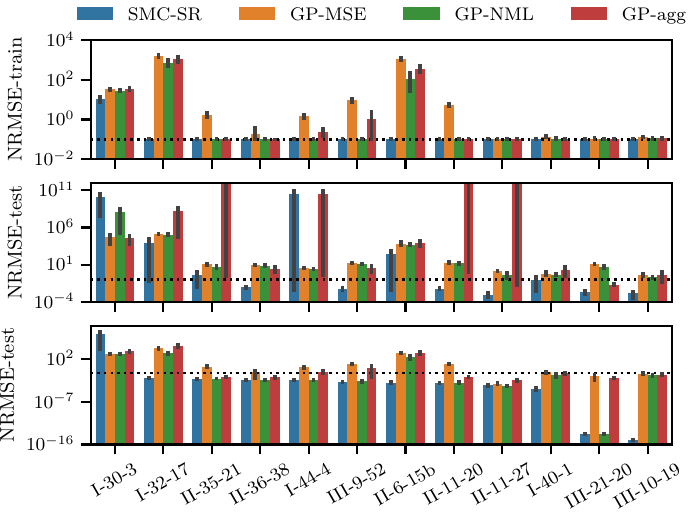}
    \caption{Results on Feynman datasets. (top) NRMSE-train for best models in final population. (middle) NRMSE-test for the same models as top. (bottom) Minimal values of NRMSE-test in final population. The black dotted line indicates the level of noise of 0.1 added to the training data.  Note: some GP-agg bars are truncated in the middle plot.}
    \label{fig:feynAgg}
\end{figure}

\begin{figure}
    \centering
    \includegraphics{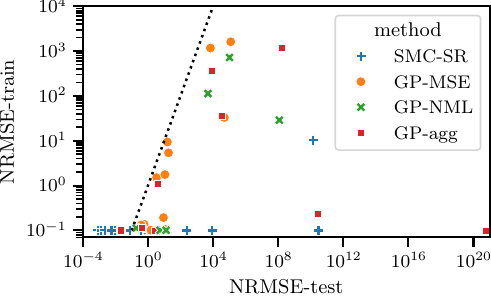}
    \caption{Overfitting on Feynman datasets. The black dotted line indicates equal NRMSE-train and NRMSE-test, overfitting is indicated to the right of this line.}
    \label{fig:feynOverfitting}
\end{figure}

Looking at the results for all the datasets (Figure \ref{fig:feynAgg}) elucidates more insights into the performance behaviours of the SMC algorithm.  The SMC algorithm is much less likely to underfit the datasets, \emph{i.e.}, have a NRMSE-train greater than the noise level, see the top of Figure \ref{fig:feynAgg}.  The middle of Figure \ref{fig:feynAgg} illustrates the NRMSE-test of the expressions with best training loss (NML in SMC-SR and explicitly defined above for GPSR methods).  The bars in the plot that land above their counterparts in the top of Figure \ref{fig:feynAgg} indicate overfitting. Figure \ref{fig:feynOverfitting} compares these NRMSE-train and NRMSE-test directly.  Overfitting is seen in almost all datasets for the GPSR approaches. SMC-SR, however, only overfits on 5 of the 12 datasets, which indicates the potential for improved generalizability with the SMC-SR.  The bottom of Figure \ref{fig:feynAgg} illustrates the minimum NRMSE-test of the final populations, again highlighting the ability to encounter highly-fit expressions.  SMC-SR has most frequent and lowest values in this figure which indicates that it is more likely to encounter these highly-fit expressions. 

\begin{figure}
    \centering
    \includegraphics{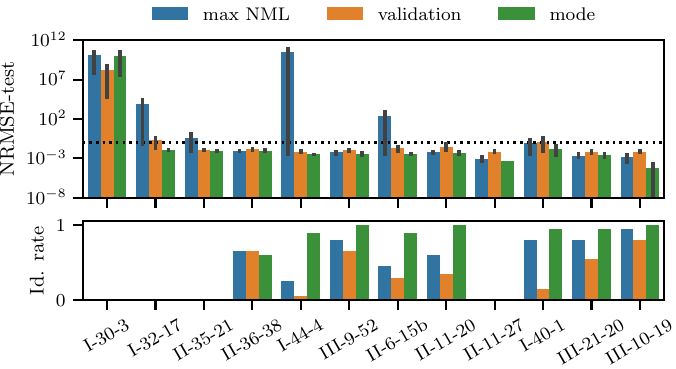}
    \caption{Effect of model selection methodology on (top) NRMSE-test and (bottom) ground-truth identification rate for the Feynman datasets.}
    \label{fig:modelSelection}
\end{figure}

The middle plot in Figure \ref{fig:feynAgg} indicates performance using standard best-fit model selection while the bottom plot indicates the performance that can be achieved with optimal model selection. Comparison of the these plots indicate that accurate model selection can significantly improve SR results, especially in cases with large pools of expressions to select from. Note that the optimal model selection in Figure \ref{fig:feynAgg} (bottom) is identified using oracle test set information which is unavailable in practice. Here, we investigate three realistic manners of model selection with respect to SMC-SR: (i) the expression with best training error, which is an approximation of the MAP expression given an uninformative prior, (ii) the expression which has the best error on a validation dataset, 20\% the size of the training dataset, and (iii) the mode of the expressions in the final population $\mathcal{P}$.  Here the mode is another approximation of the MAP expression, but without assumption on the prior.  The top of Figure \ref{fig:modelSelection} illustrates that both validation (validation) and selection of the mode (mode) are improvements over model selection by best training loss (max NML).  Interestingly, we see that selection of the mode performs better than validation selection despite the lack of additional data.  The same trend is present when considering the ability to correctly identify the ground-truth expression, as seen in Figure \ref{fig:modelSelection}.  Here ground-truth identification rate is quantified by refitting numerical constants $\pmb{\theta}$ in a given expression using the test dataset and identifying if the refit NRMSE-test is less than 1e-10.  Comparing the top of Figure \ref{fig:modelSelection} with the bottom plot in Figure \ref{fig:feynAgg} illustrates that SMC-SR using mode model selection performs near optimally in many datasets and typically better than the most optimistic cases for the GPSR methods.  

\section{Discussion}
\label{sec:discussion}

The four major differences between SMC-SR and GPSR are the use of NML, resampling, probabilistic selection, and likelihood-tempering. Comparisons of GP-NML and GP-MSE in the previous section indicate that the use of NMLL contributes modestly to the success of SMC-SR.  Resampling increases selection pressure, but the poor performance of GP-agg indicates that this alone is insufficient for successful SR.  We theorize that the remaining two differences, probabilistic selection and likelihood-tempering (either by themselves or in conjunction with the other two) are responsible for the majority of the success of SMC-SR.

\begin{figure}
    \centering
    \captionsetup[subfigure]{belowskip=-1pt}
    \begin{subfigure}[b]{0.49\textwidth}
        \centering
        \includegraphics{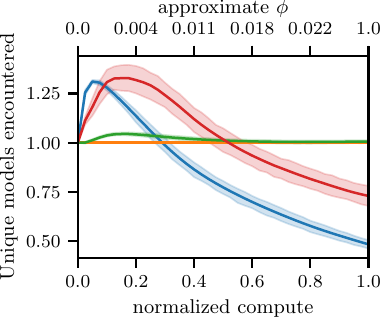}
        \caption{}
        \label{fig:totalUnique}
    \end{subfigure}
    \hfill
    \begin{subfigure}[b]{0.49\textwidth}
        \centering
        \includegraphics{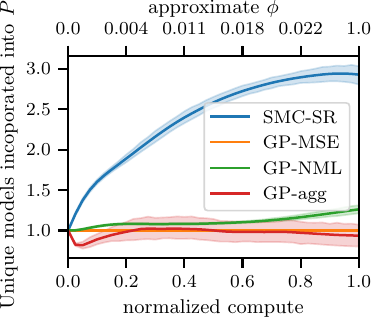}
        \caption{}
        \label{fig:totalUniqueInPop}
    \end{subfigure}
    \caption{Diversity metrics for Feynman dataset I-32-17. (a) Total unique models encountered and (b) unique models encountered that are \emph{accepted} in the MCMC kernel and are incorporated into $\mathcal{P}$. Both are normalized by GP-MSE.  Note: models that are not accepted typically have low posterior density.}
\end{figure}

It has been found that GPSR tends to struggle in efficiently exploring the domain of expressions and tends to revisit the same expressions frequently \cite{kronbergerInefficiencyGeneticProgramming2024}.  This has given rise to a few techniques that increase emphasis on novelty of expressions in SR \cite{bartlettExhaustiveSymbolicRegression2024,francaImprovingGeneticProgramming2025a}.  Probabilistic selection and likelihood-tempering both promote novelty and allow for less restricted exploration of expressions. However, in tracking the total number of unique expressions encountered, we find that SMC-SR usually has about 50\% of the number of unique expressions encountered compared to the GP-based methods (See Figure \ref{fig:totalUnique} for an example).  While this result could stem in part from the imperfect metric,\footnote{$\mathcal{M} = \theta_0(1+\theta_1)$ is identified as a different expression than $\mathcal{M} = (\theta_0 + \theta_1)$ despite the fact that they are the same expression given optimal values of the constants.} we posit that pure novelty alone is not what benefits SR.  Rather, novelty only in the region of high posterior probability is what provides benefit.

Comparing Figure \ref{fig:totalUnique} to Figure \ref{fig:totalUniqueInPop} illustrates that, despite fewer total unique models, SMC-SR has a much more dynamic population and accepts more unique models into $\mathcal{P}$.  Having more unique models incorporated into $\mathcal{P}$ is beneficial because (1) it indicates adaptation rather than stagnation of the proposal distribution $h(\mathcal{P}'|\mathcal{P})$ and (2) new (likely higher-posterior-density) models are being actively discovered.
Combining likelihood-tempering with probabilistic selection thus, provides an effective mechanism for \emph{targeted} novelty. It does not, however, ensure that the same expressions are not revisited: restricting the population in such a manner would mean the produced $\mathcal{P}$ would no longer represent the posterior. Though, in cases where the full posterior is not important, the combination of likelihood-tempering and/or probabilistic selection with the novelty methods cited earlier could prove fruitful.

\begin{figure}
    \centering
    \begin{subfigure}[b]{0.49\textwidth}
        \centering
        \includegraphics[width=\textwidth]{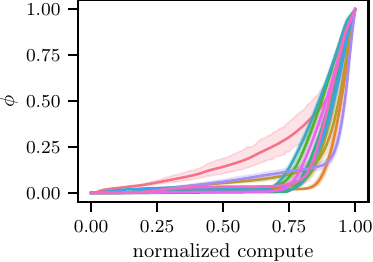}
        \caption{}
        \label{fig:feynAggPhi}
    \end{subfigure}
    \hfill
    \begin{subfigure}[b]{0.49\textwidth}
        \centering
        \includegraphics{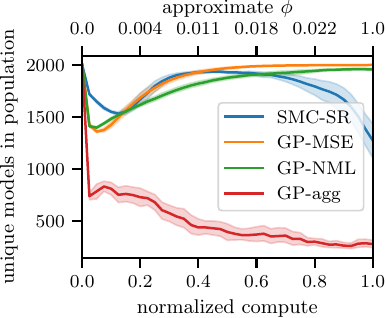}
        \caption{}
        \label{fig:popDiv}
    \end{subfigure}
    \caption{Exploration-exploitation trends of SMC-SR. (a) Adaptively set $\phi$ for Feynman datasets; each colour is a different dataset. (b) Population diversity for dataset I-32-17.}
\end{figure}

The rates of the adaptive likelihood-tempering are illustrated in Figure \ref{fig:feynAggPhi}, where we can see that the typical progression of $\phi_t$ proceeds through two phases, an exploration phase with $\phi_t \ll 1$ followed by an exploitation phase where $\phi_t$ rapidly approaches $1$.  The population dynamics illustrate a similar story in Figure \ref{fig:popDiv}, wherein the number of unique models is larger for SMC than GP early on (exploration) but then flips near the end of computation (exploitation).  In the exploitation phase, one of the benefits of duplicated expressions is that each expression can have an attempt at finding optimal $\pmb{\theta}^*$ values as part of the NML calculation.  In our work, the optimization of $\pmb{\theta}$ occurs once for each occurrence of an expression and is randomly initialized.  In future works, storing $\pmb{\theta}$ values with the expression and using them for future optimizations could add additional efficiency \cite{burlacuRevisitingGradientBasedLocal2025}.

\section{Conclusions}
\label{sec:conclusion}

This work introduced a novel Sequential Monte Carlo framework for Bayesian symbolic regression (dubbed SMC-SR) designed to enhance robustness to noise and provide built-in quantification of model-form uncertainty. Addressing the limitations of existing MCMC-based approaches and expanding on previous Bayesian GPSR efforts, SMC-SR aims to approximate the Bayesian posterior distribution over symbolic expressions. The results demonstrate that SMC-SR outperforms traditional GPSR baselines, particularly in noisy environments, exhibiting a reduced propensity for overfitting and an improved ability to identify highly-fit expressions. Furthermore, the method provides a means to quantify uncertainty in both predictions and equation form, offering a more reliable assessment of the produced expressions.

Our experiments on a challenging subset of the Feynman benchmark datasets revealed that SMC-SR achieves lower training errors more rapidly and, critically, generalizes better to unseen data. Our analysis suggests that the success of SMC-SR is attributable not only to the use of NML and increased selection pressure via resampling, but also to the combination of probabilistic selection and likelihood-tempering. These elements promote \emph{targeted} novelty and exploration of the search space, enabling the algorithm to efficiently identify regions of high posterior probability without being hampered by premature convergence or the inefficiency of pure novelty-seeking.

Since the choice of prior distributions over symbolic expressions has been shown to play a crucial role in Bayesian SR, future work could explore the impact of different priors -- particularly those informed by domain knowledge -- on the algorithm's performance and the interpretability of the resulting models.  Additionally, the incorporation of more accurate methods for calculating NML, such as replacing the Laplace approximation with SMC in the inner loop, should be investigated.


\bibliographystyle{RS}
\bibliography{refs}

\end{document}